\documentclass{article}

\PassOptionsToPackage{numbers,sort&compress}{natbib}

\usepackage{PRIMEarxiv}

\usepackage[utf8]{inputenc} 
\usepackage[T1]{fontenc}    
\usepackage{url}            
\usepackage{booktabs}       
\usepackage{amsmath,amssymb,amsfonts}
\usepackage{nicefrac}       
\usepackage{microtype}      
\usepackage{graphicx}       
\usepackage{xcolor}
\usepackage{enumitem}
\usepackage{array}
\usepackage[font=small,skip=5pt]{caption}
\usepackage{natbib}
\usepackage{fancyhdr}       
\usepackage{hyperref}       
\hypersetup{colorlinks=true,linkcolor=blue,citecolor=blue,urlcolor=blue}
\usepackage{cleveref}       

\graphicspath{{figures/}}   
\newcommand{\Pmule}{P(\mathrm{mule})}
\newcommand{\seedtime}{t_{\mathrm{seed}}}

\title{Money-Mule Detection as a Relational Learning Problem: A Study on the RBIH NFPC Dataset
}

\author{
  Owais Ahmad \\
  Applied Research, Quantiphi \\
  Marlborough, MA 01752, USA \\
  \texttt{owais.ahmad@quantiphi.com} \\
  \And
  Anirudh Deodhar \\
  Applied Research, Quantiphi \\
  Marlborough, MA 01752, USA \\
  \texttt{anirudh.deodhar@quantiphi.com} \\
}

\date{}

\begin{document}
\maketitle
\begin{abstract}
Money-mule networks move criminal funds through many small transfers, each of which looks ordinary on its own. What marks an account as a mule lies outside any single transfer: in the counterparties and addresses its network shares, and in the shape of its activity over time. Production systems remain largely rule-based and see neither, while most research classifies transactions rather than accounts, recovering account-level risk only by aggregating afterwards. We instead score the account directly, reading the bank's relational schema as a heterogeneous graph in which rows are nodes and foreign keys are edges. Each account is scored from a multi-hop neighbourhood using per-column encoders, a relative-time encoding and HeteroGraphSAGE message passing, with that neighbourhood restricted to rows predating the account's own pre-flag cutoff so that no model sees evidence recorded only after the account was flagged. On a five-year Reserve Bank Innovation Hub (RBIH) retail-banking dataset of 400 million transactions and 94{,}063 labelled accounts, 2.6\% of them confirmed mules, we compare against RelGNN, XGBoost, LightGBM and CatBoost trained on identical labels, splits and cutoffs. Our detector recovers nine in ten held-out mules at 83\% precision and a 0.5\% false-alarm rate, finding 115 more mules than the strongest tree ensemble given the same evidence. At the tuned thresholds the gap is mostly in recall: XGBoost and LightGBM lead across the first hundred alerts, but only the graph models hold their precision deep into the queue. Since the graph is the schema itself and every score is dated per account, the approach deploys on a bank's existing database without a custom feature graph build.
\end{abstract}


\section{Introduction}\label{sec:intro}
\paragraph{Context.}
Money laundering is what lets criminals use the proceeds of crime, and it operates at scale: the
United Nations Office on Drugs and Crime puts the sums cycled through it at two to five percent of
world output every year (between USD $0.8$ and $2$ trillion), of which less than one percent is
ever seized or frozen~\citep{unodc2024ml}. Almost none of that money moves in a single transaction.
It is broken into pieces, walked through accounts that appear ordinary, and reassembled somewhere.
The account in the middle belongs to a \emph{money mule}: a retail customer who receives funds they
did not earn and forwards them or withdraws them as cash, some are paid a cut, many are recruited
through a job advertisement, a relationship, or a message promising commission for a favor while few
never learn what their account was used for.

Digital payments have made that account easier to use and harder to catch. India's Unified
Payments Interface settled roughly $228$ billion transactions in $2025$~\citep{npci2026stats}, an
average of $19$ billion a month. Layering that once waited on clearing cycles now finishes in
seconds, through accounts that are cheap to open and disposable once burned. India's response has
been institutional as much as analytical: the Indian Cyber Crime Coordination Centre's Suspect
Registry has circulated more than two million first-layer mule accounts to participating
institutions~\citep{i4c2025suspect}, and the Reserve Bank has urged banks to adopt AI
models~\citep{rbi2024mulehunter}. To support that effort, the Reserve Bank Innovation Hub (RBIH)
has publicly released five years of retail banking activity, comprising about $400$ million
transactions and $96{,}091$ labelled accounts, $2{,}683$ of them confirmed mules.

\paragraph{Motivation.}
Operational monitoring remains largely a matter of rules and thresholds; a systematic review of
network analytics for anti-money laundering finds expert-written rules and hand-built features to be
the norm rather than the exception~\citep{deprez2025network}. Two limits of that arrangement are
structural, and neither is repaired by substituting a stronger classifier. Rules fire on transfers,
whereas the decision a fraud programme actually makes concerns an account: which customers are worth
an investigator's day, and whose access should be suspended. A mule's transfers, taken one at a
time, are indistinguishable from normal activity: a salary-sized credit, a same-day debit, a cash
withdrawal. Often no single transfer is anomalous at all.

What distinguishes a mule lies beyond any individual transfer, in two places. The first is
relational: mules rarely operate alone, and a recruited network leaves fingerprints in the
counterparties, addresses and postal codes that its accounts share. The second is temporal: an
account's history has a shape (a quiet stretch, a burst of inflow, a matching burst outward, silence
again), and that shape says more than any single transfer within it. Summarising each account as a
fixed set of aggregate statistics destroys both signals: the links to related accounts disappear
entirely, and the temporal ordering collapses into means and counts. This is the case for modelling
the data as a graph and retaining every relational and structural feature.

\paragraph{Our approach.}
Rather than designing a graph, we read the one the database already defines. Following relational
deep learning~\citep{fey2024rdl,robinson2024relbench}, every row becomes a node typed by its table,
every primary--foreign key reference becomes an edge in both directions, and every timed row retains
its own timestamp. The schema itself therefore supplies the graph, and no bespoke construction step
is needed beyond assembling the released tables. Each account is then scored from a temporal sample
of its own neighbourhood, taken as of that account's scoring time, and the resulting embedding is
read out as a mule probability that a tuned threshold converts into an alert. 

\begin{figure}[t]
  \centering
  \includegraphics[width=0.92\textwidth]{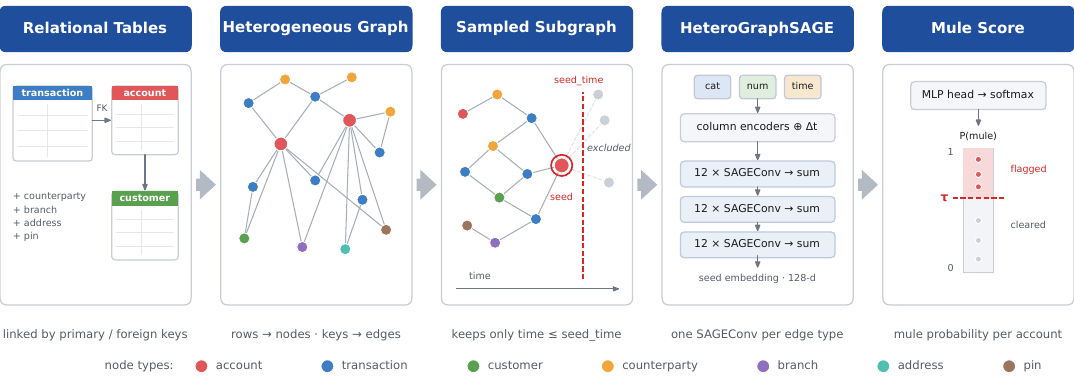}
    \caption{Overview of the pipeline, left to right. The released relational tables, linked by
    primary--foreign keys, are converted into a heterogeneous graph in which each row becomes a node
    typed by its table and each key reference becomes an edge. The seven node types are account,
    transaction, customer, counterparty, branch, address and pin. For each account to be scored (the
    seed), a subgraph is sampled around it that keeps only nodes timestamped at or before the seed
    time; later nodes are excluded. Column encoders embed each node's categorical, numerical and time
    attributes, combined with a relative time encoding $\Delta t$. A HeteroGraphSAGE network then
    propagates messages through three layers, each applying one SAGEConv per edge type (12 in total)
    and summing the results. The resulting 128-dimensional seed embedding is passed to an MLP head with
    softmax, which outputs the account's mule probability; a tuned threshold $\tau$ then flags or
    clears the account.}
  \label{fig:pipeline}
\end{figure}

\section{Related Work}\label{sec:related}

The default detector for tabular risk has long been the gradient-boosted decision
tree~\citep{chen2016xgboost,ke2017lightgbm,prokhorenkova2018catboost}, and on tabular data these
ensembles remain competitive with deep alternatives~\citep{grinsztajn2022tabular}. The concerns that
accompany them are equally long-standing, chief among them leakage introduced when a feature is
computed over all labels rather than over training labels alone~\citep{kaufman2012leakage}. Their
persistence is not only inertia: GADBench later showed that tree ensembles supplied with simple
neighbourhood aggregates match purpose-built graph detectors across ten
datasets~\citep{tang2023gadbench}, which is why gradient boosting appears below as a serious
baseline rather than a formality.

In parallel, the graph literature developed its own vocabulary for aggregating over a node's
neighbourhood: sampled mean-and-concatenate updates~\citep{hamilton2017graphsage}, attention over
neighbours~\citep{velickovic2018gat,brody2022gatv2}, and per-relation parameterisations for typed
graphs~\citep{schlichtkrull2018rgcn,hu2020hgt}. Financial applications followed. Transaction
classification on labelled blockchain data~\citep{weber2019elliptic} was joined by aggregators
designed to resist camouflaged neighbourhoods~\citep{dou2020caregnn,liu2021pick}, by risk-aware
propagation schemes~\citep{tang2022bwgnn,xiang2025enhancing}, and by laundering detection at
production scale~\citep{cardoso2022laundrograph,egressy2024provably}. More recent operators read a
node's time-ordered edge sequence directly~\citep{verlaan2025lasgnn} or bias attention using edge
attributes~\citep{lin2024fraudgt}. Two properties of this work shaped our design. First, most of it
operates at the transaction or edge level and implicates the account only afterwards, through
aggregation, whereas our labels attach directly to account holders. Second, its public datasets are
synthetic, drawn from cryptocurrency, or concerned with a different financial
crime~\citep{weber2019elliptic,huang2022dgraph,altman2023amlworld}.

Work aimed specifically at mule \emph{accounts} arrived later, is limited in volume, and relies
throughout on proprietary data. DELATOR~\citep{assumpcao2022delator} learns client-level embeddings
on a Brazilian bank's temporal transaction graph, but does not condition each client on its own
as-of date. The model architecturally closest to ours, \citet{johannessen2025dnb}, trains a
heterogeneous message-passing network on a Norwegian bank's graph of customers and counterparties at
well under one percent prevalence. It classifies nodes transductively over the full graph, splitting
only the labels. We also keep one shared graph, but score every account only from its own
cutoff-bounded neighbourhood. MuleTrack~\citep{jambhrunkar2025muletrack} takes a behavioural route,
fitting a lightweight Markov model over hand-specified weekly states, which we borrow only to
characterise accounts (\Cref{fig:states}). \citet{zhang2026mule} detect mule customers with
gradient-boosted trees under a single global temporal split.

Two of these settings locate the exposure in the evaluation protocol rather than in any individual
feature: transductive inference~\citep{johannessen2025dnb} and a global temporal
split~\citep{zhang2026mule,egressy2024provably} both give every account the same cutoff regardless
of when its label was decided. A related question concerns what is reported. Because analyst
capacity, not accuracy, is the operational constraint, we report queue-based metrics alongside
threshold-free ones.

The framing we adopt comes from a separate line of work. Relational deep learning treats a database
as a graph already: rows are nodes, foreign-key references are edges, and a task is defined per
entity at a given timestamp~\citep{fey2024rdl,robinson2024relbench}. The graph then needs no bespoke
construction step, and temporal neighbour sampling provides a single auditable point at which to
enforce a per-account cutoff. Within this setting, RelGNN composes messages along atomic routes
derived from the schema and reports the strongest current results among relational graph
models~\citep{chen2025relgnn}; we run it as our second graph model.

\section{Problem Formulation}\label{sec:problem}
\paragraph{The database as a graph.}
Following \citet{fey2024rdl}, we read the institution's relational database directly as a
heterogeneous graph $G=(V,E,\tau)$. Every row becomes a node, typed by the table it comes from; every
primary--foreign key reference becomes an edge, materialised in both directions; and $\tau(v)$ is the
timestamp of the row behind node $v$. Some rows have no time of their own, a shared identifier, for
instance and for these we set $\tau(v)=\bot$.

\paragraph{Labels.}
Let $\mathcal{A}\subset V$ be the set of account nodes and $\mathcal{L}\subseteq\mathcal{A}$ the
labelled subset, where $y_i\in\{0,1\}$ records whether account $a_i$ was confirmed to be a money
mule. Each positive account also carries a \emph{flag date} $f_i$, the date on which the
investigation concluded.

\paragraph{Scoring cutoffs.}
The flag date is an output of an investigation, so anything recorded as a result of that
investigation falls on or after $f_i$. A model that claims to detect the account \emph{beforehand} must not see
any of it. We therefore give every labelled account its own scoring cutoff,
\begin{equation}
\seedtime(i) =
\begin{cases}
\min\!\big(f_i - \Delta,\; T_{\mathrm{end}}\big), & y_i = 1,\\[3pt]
\rho(a_i), & y_i = 0,
\end{cases}
\label{eq:seedtime}
\end{equation}
where $\Delta$ is one day, $T_{\mathrm{end}}$ is the end of the modelling window, and
$\rho(a_i)=t^{\mathrm{last}}_i-\Delta$ with $t^{\mathrm{last}}_i$ the date of the account's last
transaction. A mule is thus scored on its history up to one day before its flag date, and a non-mule
on its history up to one day before its last transaction; non-mule accounts with no transactions are
dropped. The positive
case is dictated by the data. The negative case is not: a non-mule account carries no flag date and hence
no natural as of date, so the cutoff rule $\rho$ is a modelling choice.

\paragraph{Admissible evidence.}
Given a cutoff, the evidence admissible for $a_i$ is whatever existed in the graph at that time:
\begin{equation}
V_i \;=\; \big\{\, v \in V \;:\; \tau(v) = \bot \ \text{ or } \ \tau(v) \le \seedtime(i) \,\big\},
\label{eq:admissible}
\end{equation}
and $G_i$ is the $L$-hop neighbourhood of $a_i$ in the subgraph induced by $V_i$. Untimestamped nodes
are admissible because they carry no time-dependent information: a shared counterparty, address or
PIN is only a routing device, branch attributes carry no timestamp and are treated as static, and every
timestamped attribute reached through them is itself filtered by
\Cref{eq:admissible}.

A model is then any function $f_\theta$ that depends on $a_i$ only through $G_i$,
\begin{equation}
s_i \;=\; f_\theta(G_i) \;\approx\; \Pr\!\big(y_i = 1 \mid G_i\big),
\label{eq:model}
\end{equation}
and we impose \Crefrange{eq:admissible}{eq:model} on \emph{every} model we report, graph or tabular.
Two properties follow. Each account is cut at its own date, so this is not a global temporal split.
Nothing outside $G_i$ enters the computation, so scoring is inductive.

\paragraph{Data and metrics.}
\Crefrange{eq:seedtime}{eq:model} are stated for any labelled entity table in any
schema; what follows binds them to our dataset, which carries labels for $96{,}091$ accounts. Of these,
$\vert\mathcal{L}\vert = 94{,}063$ admit a well-defined cutoff under \Cref{eq:seedtime}, and $2{,}488$
are mules---a prevalence of $2.6\%$. At that rate a model that answers ``not a mule'' everywhere is
$97.4\%$ accurate, so accuracy carries no information. We report ROC-AUC and PR-AUC for ranking
quality, together with the queue quantities an analyst actually experiences,
\begin{equation}
\mathrm{P}@k = \frac{1}{k}\sum_{i \in \mathcal{S}_k} y_i,
\qquad
\mathrm{R}@k = \frac{\sum_{i \in \mathcal{S}_k} y_i}{\sum_{i} y_i},
\end{equation}
where $\mathcal{S}_k$ is the set of the $k$ highest-scoring held-out accounts. At a fixed operating
point we also report precision, recall, $F_1$ and the false-alarm rate, along with the confusion
counts behind them. Every threshold is selected on validation data and applied unchanged; whether it
survives out of sample turns out to be one of the more informative differences between the models
(\Cref{sec:ranking}). Throughout, the unit of prediction is the account, not the ring.

\begin{table}[t]
\centering
\caption{The seven tables of the assembled RBIH database, the node type each becomes, and the columns
each contributes as features. Hub tables are deliberately featureless: they carry connectivity only.
Row counts are after restriction to labelled accounts and their transactions.}
\label{tab:schema}
\small
\begin{tabular}{@{}llrl@{}}
\toprule
Table & Node type & Rows & Feature columns \\
\midrule
\texttt{transaction}  & \textsc{transaction}  & 237{,}785{,}028 & 5 categorical, 2 numerical, timestamp \\
\texttt{counterparty} & \textsc{counterparty} & 105{,}107 & --- (hub) \\
\texttt{account}      & \textsc{account}      & 96{,}091  & 9 categorical, opening date \\
\texttt{customer}     & \textsc{customer}     & 95{,}820  & 12 categorical, 2 dates \\
\texttt{address}      & \textsc{address}      & 69{,}616  & --- (hub) \\
\texttt{branch}       & \textsc{branch}       & 9{,}000   & 3 categorical, 3 numerical \\
\texttt{pin}          & \textsc{pin}          & 5{,}720   & --- (hub) \\
\bottomrule
\end{tabular}
\end{table}

\section{Dataset}\label{sec:data}
\begin{figure}[h]
  \centering
  \includegraphics[width=0.74\textwidth]{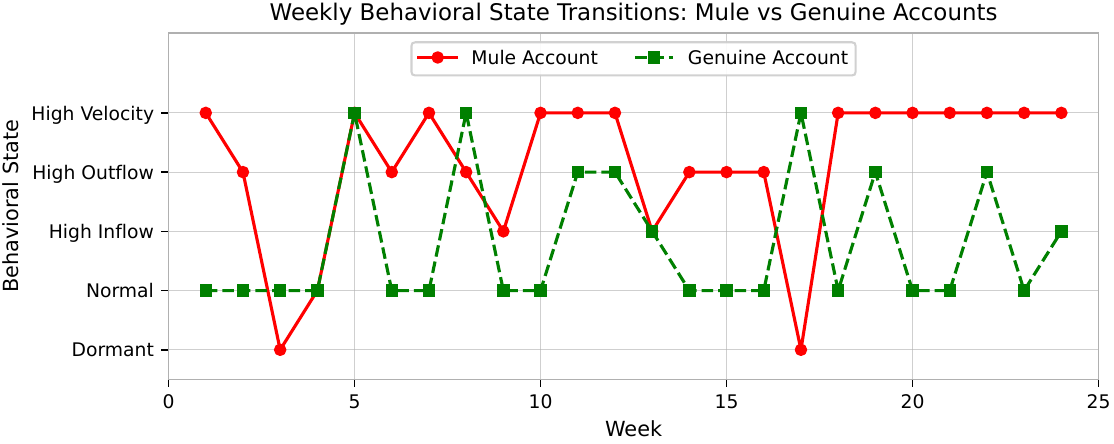}
  \caption{Weekly behavioural state of one flagged mule and one genuine account over the 24 weeks
  before their cutoffs. States come from each account's own pre-cutoff transactions via population
  deciles of weekly relative velocity and credit share; none is derived from labels. The mule
  alternates high-velocity and directional-flow weeks, with brief dormant spells, and ends in a
  sustained high-velocity run; the genuine account returns to
  \textsc{normal} between occasional excursions.}
  \label{fig:states}
\end{figure}

The data comes from the Reserve Bank Innovation Hub (RBIH)\footnote{\url{https://www.kaggle.com/datasets/abhyudayrbih/rbih-nfpc-phase-2/data}} and covers five years of retail banking
activity at Indian institutions, from 2020-07-01 to 2025-07-12. It comprises the seven tables of
\Cref{tab:schema}, and is restricted to the labelled accounts and their transactions, which reduces the
roughly $400$ million released transactions to the $237.8$ million rows of the transaction table. Columns recording the aftermath of an
investigation---account status, freeze and unfreeze dates, alert reason, flagging branch---are removed
when the database is written, as are reporting-time balances and the IP field, which at $66\%$ null
and nearly unique per transaction carries identifiers rather than structure. Of the $96{,}091$
labelled accounts ($2{,}683$ mules), $94{,}063$ admit a well-defined cutoff under \Cref{eq:seedtime}.
The rest are $46$ mules flagged before the window opens, $1{,}809$ non-mules with no transactions, and
$173$ accounts ($149$ of them mules) whose cutoff precedes account opening. This leaves $2{,}488$
mules, a prevalence of $2.6\%$. A class-stratified $70/15/15$ holdout with a fixed seed separates
entities rather than time, temporal safety being carried per account by the cutoff. Visible history
differs by class---mules have a median of $1{,}196$ transactions against $1{,}987$ for non-mules---and
so does its shape: \Cref{fig:states} shows the weekly trajectory of a mule and a genuine account over
the 24 weeks before their cutoffs, the mule spending most weeks in high-velocity or
high-outflow states while the genuine account mostly stays normal.

\section{Method}\label{sec:method}

\subsection{From relational schema to heterogeneous graph}\label{sec:graph}
\begin{figure}[h]
  \centering
  \includegraphics[width=\textwidth]{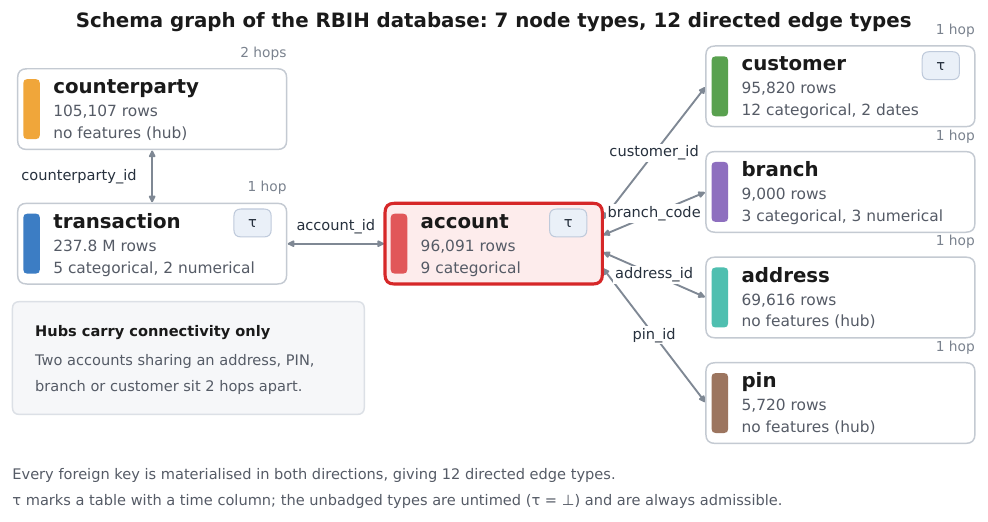}
  \caption{The heterogeneous graph that the schema of \Cref{tab:schema} yields under
  the rule above---read off the database rather than designed. Every row is a node typed by its
  table and every primary--foreign key reference is an edge materialised in both directions, giving
  twelve directed edge types over seven node types. A $\tau$ badge marks a table carrying a time
  column; the unbadged types are untimed, $\tau(v)=\bot$, and so are always admissible under
  \Cref{eq:admissible}. \textsc{Counterparty}, \textsc{address} and \textsc{pin} hold no features
  of their own: they exist as shared endpoints, which places accounts that share one of
  them within a short path of each other without a clique between them.}
  \label{fig:graph}
\end{figure}

We read the graph directly off the assembled schema. Six foreign-key references---transaction
to account and to counterparty, and account to customer, branch, address and PIN, the last two carried
onto the account from its customer record when the database is assembled---connect the seven node
types of \Cref{tab:schema}. Each is materialised in both directions, so that messages travel from a
transaction up to the account that owns it and back down~\citep{fey2024rdl}, giving twelve directed
edge types. The rule doing this work is schema-agnostic---one node type per table,
one bidirectional edge type per foreign key---so the graph an institution gets is whatever its own
schema already encodes. \Cref{fig:graph} draws the graph the instance evaluated here yields: which
tables become which node types, which references connect them, and which of those types carry a
timestamp for the cutoff of \Cref{eq:admissible} to act on.

Node features are the table's own columns, typed once and left alone. \textsc{Counterparty},
\textsc{address} and \textsc{pin} carry none: they exist purely as shared endpoints, so two accounts
sharing an address or PIN code sit two hops apart, and two accounts paying the same counterparty sit
four hops apart, in each case without any clique between them. This keeps the graph linear rather than quadratic in the number of accounts sharing an
identifier, and it is where the fanout budget of \Cref{sec:sampling} is spent. Relational context must therefore be obtained by message passing, which
leaves the cutoff of \Cref{eq:admissible} as the only point at which post-cutoff information could enter.

\subsection{Leakage-safe temporal neighbour sampling}\label{sec:sampling}
\begin{figure}[h!]
  \centering
  \includegraphics[width=\textwidth]{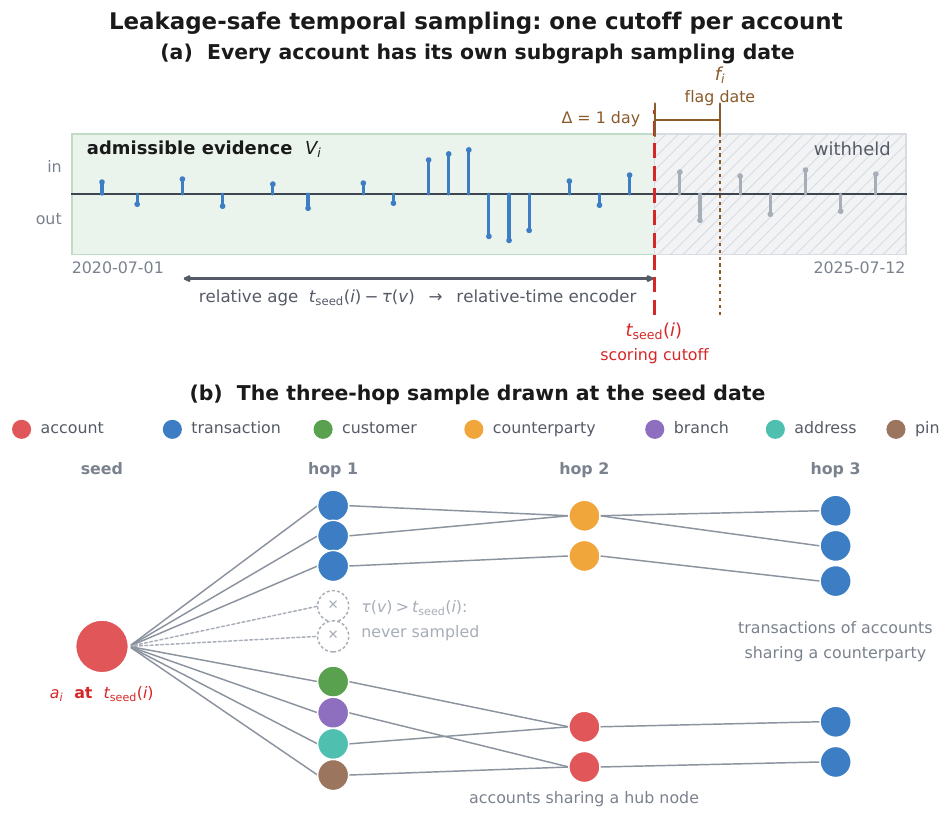}
  \caption{Leakage-safe temporal neighbour sampling. \textbf{(a)} One seed account's
  transactions on a time axis, credits above the line and debits below. Rows at or before that
  account's own cutoff $\seedtime(i)$ make up the admissible evidence $V_i$ of
  \Cref{eq:admissible}; the hatched region, running through the flag date
  $f_i$ and beyond, is withheld from every model we report; $f_i=\seedtime(i)+\Delta$ unless
  $f_i-\Delta$ exceeds $T_{\mathrm{end}}$. The arrow beneath
  the axis is the quantity the relative-time encoder of \Cref{sec:encoders} consumes: a row's age
  measured against that account's own cutoff rather than against any absolute date. \textbf{(b)} The
  three-hop sample drawn at that date. Hollow dashed nodes are candidates whose timestamp fails the
  cutoff; they never enter the subgraph, so nothing reachable only through them enters either.
  Untimed rows have no timestamp to test and are admitted unconditionally.}
  \label{fig:sampling}
\end{figure}

For the graph models, the cutoff is enforced in exactly one place: the sampler. For a batch of seed accounts we draw a
three-hop neighbourhood with a temporal neighbour loader~\citep{fey2019pyg}, configured with each
seed's own $\seedtime(i)$ as its input time. A candidate node joins the subgraph only if its timestamp
satisfies \Cref{eq:admissible}, so one graph sits on disk with no per-split copies and a single place
to audit. The per-hop fanout is $[4096, 512, 128]$. The wide first hop is forced rather than chosen:
with a median of $1{,}987$ visible transactions per non-mule, a small hop-one budget would subsample
precisely the history the task is about. At batch size $192$ this admits up to $786{,}432$ hop-one
transaction nodes per step; training runs $343$ steps per epoch and peaks at $51.88$\,GiB on an
80\,GB A100. Sampling is stochastic and we do not average over draws, so a fresh pass differs
slightly from the training-time scores. \Cref{fig:sampling} shows the two halves of
this together: the per-account cutoff that decides what counts as admissible, and the three-hop
sample the loader returns once that cutoff has been applied.

\subsection{Column encoders and relative-time encoding}\label{sec:encoders}
Every node type is encoded into a shared $128$-dimensional space by a per-column encoder
bank~\citep{robinson2024relbench}: embeddings for categorical fields, a linear map for numerical ones,
and cyclical features for timestamps, followed by a four-layer ResNet table encoder per node type.
Two departures from the default recipe follow from wanting a
model that can score tomorrow's rows without re-reading yesterday's.

First, rather than standardising each numerical column by its dataset mean and standard deviation, we
squash it row-locally inside the forward pass with
$x \mapsto \operatorname{sign}(x)\log(1+\lvert x\rvert)$. A value therefore maps to the same number in
any future batch, and the transform travels with the checkpoint. For consistency we also neutralise
the dataset-derived statistics the encoders would otherwise use---the mean and standard deviation on
the eight numerical columns (the five of \Cref{tab:schema} plus a constant placeholder on each
featureless hub), and the year range on the four timestamp columns---retaining only the
categorical cardinalities.

Second, a relative-time encoder maps $\seedtime(i) - \tau(v)$, in days, to a vector added to every
timestamped node's embedding. The model therefore sees each row's age with respect to the account's
own cutoff, in addition to the calendar features of the timestamp encoder, which is what makes a
per-account cutoff meaningful at all.
It is also the channel through which any regularity in how cutoffs were assigned would reach the
model.

\subsection{Message passing, readout and training}\label{sec:loss}
The message-passing block is deliberately standard. Each of three layers holds one
\textsc{SAGEConv}~\citep{hamilton2017graphsage} per edge type---thirty-six in total---whose outputs are
summed across relations at every destination node, followed by per-node-type layer normalisation and a
ReLU. Three hops are what the task needs: an account's own transactions, address and PIN, then the
counterparties its transactions touch and the other accounts sharing its address or PIN, then the wider
activity around them. Readout passes the seed
account rows through a single linear layer to two logits, whose softmax gives
$s_i = \Pmule$. The two-class head keeps the loss an ordinary weighted cross-entropy and the score a
directly thresholdable probability.

Training up-weights the positive class by $\min(n_{\mathrm{neg}}/n_{\mathrm{pos}}, 100) = 36.80$ and
uses AdamW~\citep{loshchilov2019adamw} at learning rate $10^{-3}$ and weight decay $10^{-4}$, with
gradient-norm clipping at $1.0$. We early-stop on validation ROC-AUC with patience $20$; the run
reported below stopped at epoch $29$ having selected epoch $9$. Selecting on ROC-AUC while reporting
PR-AUC is a mismatch, and since the selected epoch was not the best-PR-AUC epoch, our PR-AUC is
pessimistic by an unmeasured amount. Because probabilities at this prevalence are not calibrated
around $0.5$, the operating point is a separate decision made on validation only: we take the
$F_1$-optimal threshold there and apply it unchanged to the held-out split. Scoring is inductive, so a
newly opened account needs no rebuild and no retraining.

\section{Experimental Setup}\label{sec:setup}
\paragraph{Baselines.}
We compare against four systems. One is a stronger graph architecture given the identical relational
input, which asks whether a more expressive message-passing scheme pays off here. The other three are
the standard tabular models, given the same underlying evidence flattened into per-account rows, which
asks how much the graph structure contributes at all. RelGNN~\citep{chen2025relgnn} is the first of
these: it replaces our message-passing block and leaves everything else in place---fourteen composite
atomic routes derived from our schema, two layers of eight heads at $128$ channels, summed across
routes. Only the batch size differs. At this fanout a batch of $192$ exhausts 80\,GB, so RelGNN is trained at $96$, a real asymmetry that should be
kept in mind when comparing the two graph models.

The three gradient-boosted models---XGBoost~\citep{chen2016xgboost}, LightGBM~\citep{ke2017lightgbm}
and CatBoost~\citep{prokhorenkova2018catboost}---receive the same evidence flattened into $58$
per-account features: static attributes of the account, its customer and its branch, three age
quantities measured against the cutoff, and twenty-eight aggregates of the account's own visible
transactions (counts, amount moments, debit/credit splits, distinct counterparties, channels, MCCs,
sub-types, recency and span). Every aggregate obeys the same $\tau(v) \le \seedtime(i)$ filter the
sampler applies, so the trees see no later history than the graph models do, and in fact all of it,
where the sampler caps hop one at $4{,}096$ transactions. What they do not see is structure. XGBoost and CatBoost run to $3{,}000$ rounds with early stopping on validation AUC
(patience $50$); LightGBM uses a fixed $400$ rounds, since on this imbalance early stopping halts it
at the first or second iteration and leaves a stump. All three are run over three seeds and selected
on validation PR-AUC. Across all five systems the label tables, split masks, cutoffs, class weighting
of $36.80$ and threshold-selection protocol are identical, and none receives a label-derived feature.

\paragraph{Implementation.}
The pipeline is built on PyTorch Geometric~\citep{fey2019pyg}, with
RelBench~\citep{robinson2024relbench} providing graph construction and the column and temporal
encoders. Each run occupies one 80\,GB A100, and the graph is cached once. Cost is dominated by neighbour
sampling: our model ran $343$ steps per epoch at $5$--$6$ minutes an epoch, while RelGNN at batch $96$
ran $686$ steps at roughly $11$ minutes an epoch and
early-stopped at epoch $93$---about three hours of wall clock end to end for ours against eighteen for
RelGNN. For the curves of \Cref{sec:ranking} the
gradient-boosted models reproduce their recorded metrics exactly, while the graph models are re-run in
inference mode, where stochastic sampling matches the recorded ROC-AUC and PR-AUC to within $0.003$.

\section{Results}\label{sec:results}

\subsection{Ranking quality and fixed thresholds}\label{sec:ranking}
\begin{figure}[h]
  \centering
  \includegraphics[width=0.48\textwidth]{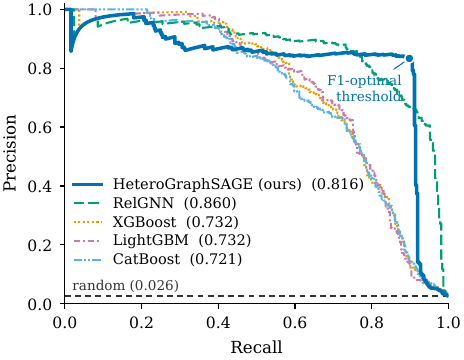}
  \caption{Precision-recall on the held-out split, PR-AUC in the legend; the dashed line is the
  prevalence-random baseline ($0.026$) and the dot our $F_1$-optimal operating point. Graph-model
  scores come from a re-inference pass, exact to within $0.003$.}
  \label{fig:pr}
\end{figure}

\Cref{tab:ranking} reports threshold-free quality on the held-out split: $373$ mules among $14{,}110$
accounts, a base rate of $0.0264$. All five models rank far above chance, so the comparisons that
matter are between them. RelGNN ranks best, exceeding our model by $0.026$ ROC-AUC and $0.044$ PR-AUC.
The three trees cluster at $0.721$--$0.733$ PR-AUC, about $0.08$--$0.09$ below our model and $0.13$ below
RelGNN, and are separated from one another by $0.011$, comparable to the seed-to-seed spread of a single tree ensemble.
\Cref{fig:pr} shows where the graph models earn the margin: the trees match or exceed them up to
roughly $0.4$ recall, and the graph models pull ahead sharply after it.

\begin{table}[t]
\centering
\caption{Threshold-free quality on the held-out split ($373$ mules / $14{,}110$ accounts, base rate
$0.0264$). Both graph models are single-seed; the gradient-boosted rows are the seed selected on
validation PR-AUC out of three. Best per column in \textbf{bold}.}
\label{tab:ranking}
\small
\begin{tabular}{@{}llcc@{}}
\toprule
Model & Family & ROC-AUC & PR-AUC \\
\midrule
HeteroGraphSAGE (ours) & GNN & 0.9592 & 0.8154 \\
RelGNN                 & GNN & \textbf{0.9848} & \textbf{0.8598} \\
XGBoost                & GBM & 0.9541 & 0.7324 \\
LightGBM               & GBM & 0.9479 & 0.7325 \\
CatBoost               & GBM & 0.9538 & 0.7213 \\
\bottomrule
\end{tabular}
\end{table}

Ranking quality shows what a model could deliver, while a fixed threshold shows what it does deliver,
and the ordering of the two graph models reverses once a threshold is fixed. \Cref{tab:operating} reports the $F_1$-optimal operating point of
\Cref{sec:loss}, selected on validation and applied unchanged. Our model attains $F_1 = 0.8663$ at
$83.2\%$ precision and $90.3\%$ recall, catching $337$ of $373$ mules against $68$ false alarms.
RelGNN, the better ranker, drops to $F_1 = 0.7930$: its validation-optimal threshold of $0.980$ sits
in the extreme tail of its score distribution, where a small validation-to-test shift costs a large
amount of recall, whereas ours sits at $0.640$ in a flatter region. The trees land between $0.654$ and
$0.680$ $F_1$, limited by recall rather than precision. The reading is not that one model is better
but that the two suit different products: RelGNN threshold-free ranking, ours a fixed-threshold
decision rule.

\begin{table}[t]
\centering
\caption{The $F_1$-optimal operating point on the held-out split. Each model's threshold is selected on
validation and applied unchanged; FA is the false-alarm rate on the $13{,}737$ non-mules. Best per
column in \textbf{bold}.}
\label{tab:operating}
\footnotesize
\begin{tabular}{@{}lrccccrrrr@{}}
\toprule
Model & Thr. & $F_1$ & P & R & FA & TP & FP & FN & TN \\
\midrule
HeteroGraphSAGE (ours) & 0.640 & \textbf{0.8663} & \textbf{0.832} & \textbf{0.903} & 0.50\% & \textbf{337} & 68 & \textbf{36} & 13{,}669 \\
RelGNN                 & 0.980 & 0.7930 & 0.795 & 0.791 & 0.55\% & 295 & 76 & 78 & 13{,}661 \\
XGBoost                & 0.905 & 0.6772 & 0.728 & 0.633 & 0.64\% & 236 & 88 & 137 & 13{,}649 \\
LightGBM               & 0.810 & 0.6799 & 0.793 & 0.595 & \textbf{0.42\%} & 222 & \textbf{58} & 151 & \textbf{13{,}679} \\
CatBoost               & 0.905 & 0.6543 & 0.700 & 0.614 & 0.71\% & 229 & 98 & 144 & 13{,}639 \\
\bottomrule
\end{tabular}
\end{table}

\subsection{Analyst-queue behaviour}\label{sec:queue}
In practice, analysts work through a queue of fixed length rather than apply a threshold. \Cref{tab:queue} therefore
reports precision and recall among the top $k$ scored accounts, and it holds the sharpest contrast in
this study. At $k=100$ the trees match or beat both graph models: XGBoost places $99$ mules in its first hundred alerts,
LightGBM $98$ and CatBoost $96$, against $92$ for our model and $96$ for RelGNN. A hundred alerts is a
plausible daily budget for a small review team, and on that budget a well-featurised tree is the
better tool. By $k=500$ the ordering has reversed and the margin is wide: the graph models hold
$0.676$--$0.680$ precision while all three trees fall to $0.562$--$0.564$. The trees are excellent at
the very top of their ranking and poor across the rest of it, which is their weak thresholded recall
in \Cref{tab:operating} seen from the operational side.

\begin{table}[t]
\centering
\caption{Queue behaviour on the held-out split: precision and recall among the top $k$ scored accounts.
Best per column in \textbf{bold}.}
\label{tab:queue}
\small
\begin{tabular}{@{}lcccc@{}}
\toprule
& \multicolumn{2}{c}{Precision@$k$} & \multicolumn{2}{c}{Recall@$k$} \\
\cmidrule(lr){2-3}\cmidrule(lr){4-5}
Model & 100 & 500 & 100 & 500 \\
\midrule
HeteroGraphSAGE (ours) & 0.920 & \textbf{0.680} & 0.247 & \textbf{0.912} \\
RelGNN                 & 0.960 & 0.676 & 0.257 & 0.906 \\
XGBoost                & \textbf{0.990} & 0.562 & \textbf{0.265} & 0.753 \\
LightGBM               & 0.980 & 0.564 & 0.263 & 0.756 \\
CatBoost               & 0.960 & 0.562 & 0.257 & 0.753 \\
\bottomrule
\end{tabular}
\end{table}

At their $F_1$-optimal points the two families nearly meet on precision---$0.700$--$0.793$ for the trees against $0.795$--$0.832$ for the graph models---and stay far apart on recall,
$0.595$--$0.633$ against $0.791$--$0.903$. The best tree, LightGBM, finds $222$ of $373$ mules and ours
finds $337$, so the $0.19$ $F_1$ gap between them comes almost entirely from those $115$ accounts. Since the
trees receive the same admissible history aggregated over the same cutoff but no neighbourhood
features, the difference most likely comes from what they cannot see: the identifiers an account shares, the activity reachable through them, and the temporal
ordering that aggregation discards.
Two things bound how much of this to credit to architecture. Widening the per-hop fanout from
$[4096,32,32]$ to $[4096,512,128]$ under an otherwise identical recipe moved our validation PR-AUC from
$0.695$ to $0.762$ (test: $0.746$ to $0.815$)---a change comparable to the entire graph-versus-tree margin, so a comparison run at a
narrow fanout would measure the fanout instead. And the trees' three-seed validation PR-AUC spreads
($0.7393$--$0.7404$, $0.7267$--$0.7369$ and $0.6857$--$0.7116$) give an informal noise floor on the
tabular side and none on the graph side, so the $0.044$ PR-AUC between the two graph models should be
read with that in mind.

\section{Limitations and Future Work}\label{sec:threats}
The labels mark confirmed mules, and every other account is treated as a negative, so an
undetected mule ring is not merely absent from the positives, it is taught to the model as
normal. Our evaluation cannot measure this, because the only ground truth
available is the same investigative record the model is fitted to---every precision figure
we report is bounded by what that process happened to reach. The response is to stop
treating unlabelled accounts as confirmed negatives: positive--unlabelled training, or an
anomaly objective over the same graph, would let unusual structure surface without a prior
label. Scoring connected structures rather than individual accounts would put the unit of
detection where the behaviour actually lives; we leave this to future work.

\section{Conclusion}\label{sec:conclusion}
The question we set out to answer was whether a faithful reading of a bank's own schema, with time
respected per account, is enough to detect mule accounts well, and our results suggest that it is. The RBIH database is read directly as a
heterogeneous graph, each account is cut at its own pre-flag date, and the sampler admits only rows
recorded at or before that cut---so nothing knowable only after an investigation reaches the model. On held-out
accounts the detector reaches ROC-AUC $0.959$ and PR-AUC $0.815$, and at its tuned threshold recovers
$337$ of $373$ mules at $83.2\%$ precision and a $0.50\%$ false-alarm rate, which is $115$ more mules
than the best tree given the same evidence. 

We draw four further conclusions. At a fixed threshold,
relational structure buys recall more than precision: the two families nearly agree on how clean
their alerts are and disagree on how
many mules they find at all. A better ranker is not automatically a better detector, as RelGNN shows by
ranking above us and still losing $0.073$ $F_1$ once a threshold is fixed. And the choice of metric
depends on the review team's capacity, since XGBoost and LightGBM lead in the first hundred alerts but all
three trees fall well behind by five hundred. Finally, the cutoff for non-mule accounts is a modelling choice
rather than a fact in the data, so it has to be stated explicitly, and is better tied to an event
independent of the account's own activity.

{\small
\bibliographystyle{unsrtnat}
\bibliography{references}
}

\end{document}